\documentclass[11pt]{article}
\usepackage{acl2018_mine}
\usepackage{times}
\usepackage{url}
\usepackage{latexsym}
\usepackage{amssymb}
\usepackage{amsmath}
\usepackage{graphicx}
\usepackage{tabulary}
\usepackage{xcolor}

\aclfinalcopy 

\title{Self-Attention-Based Message-Relevant Response Generation for Neural Conversation Model}
\author{Jonggu Kim, Doyeon Kong, Jong-Hyeok Lee \\
  Computer Science and Engineering, \\
  Pohang University of Science and Technology (POSTECH) \\
  Pohang, Republic of Korea \\
  {\tt \{jgkimi,dykong,jhlee\}@postech.ac.kr} \\}

\date{9th, Mar., 2018}

\begin{document}
\maketitle
\begin{abstract}
  Using a sequence-to-sequence framework, many neural conversation models for chit-chat succeed in naturalness of the response. Nevertheless, the neural conversation models tend to give generic responses which are not specific to given messages, and it still remains as a challenge. To alleviate the tendency, we propose a method to promote message-relevant and diverse responses for neural conversation model by using self-attention, which is time-efficient as well as effective. Furthermore, we present an investigation of why and how effective self-attention is in deep comparison with the standard dialogue generation. The experiment results show that the proposed method improves the standard dialogue generation in various evaluation metrics.
\end{abstract}

\section{Introduction}

Dialogue systems are designed to have a conversation with a user. According to the objective of conversation, dialogue systems are classified into task-oriented dialogue systems which conduct specific tasks such as booking and ordering, and non-task-oriented dialogue systems (chatbots) which are constructed for chit-chat. While components of task-oriented dialogue systems are pipelined after the components are constructed separately, chatbots are usually constructed in an end-to-end way which is similar to neural machine translation models based on the sequence-to-sequence architecture. Even though such chatbots have achieved great success in naturalness of the response like human-being, but they still have a challenge called the generic response problem. The generic response problem indicates that the produced response is not informative or specific to the given message, but generic such as ``I see.'' or ``I don't know''.

Even though much recent research on the problem has been conducted, the problem has not been easily cleared; some methods are not enough effective, the other methods are complex and time-inefficient.

In this paper, we present an empirical analysis on structural reason why sequence-to-sequence models generate such responses and provide its clues. Based on the analysis, we propose a decoding method using self-attention to promote message-relevant and diverse responses for standard sequence-to-sequence models, which does not require a new model architecture. Then, we present a variety of experimental results for verification of the proposed method. The experimental results demonstrate that the proposed method generates more interesting responses than the standard dialogue generation.

In Section \ref{sec:Related Work}, we introduce previous methods to alleviate the problem as related work. In Section \ref{sec:The Proposed Method}, we introduce our motivation and the proposed method in detail. We then show the experimental settings and results in Section \ref{sec:Experiments}. We discuss the results in detail in Section \ref{sec:Discussion}. Finally, we conclude our work in Section \ref{sec:Conclusion}.

\section{Related Work}
\label{sec:Related Work}
Non-task-oriented dialogue systems often use a framework of machine translation \cite{Ritter:EMNLP11}. Recently, the framework of neural machine translation which is a sequence-to-sequence framework based on neural networks is applied for dialogue systems. In terms of a natural response, such dialogue systems are in great success. However, in terms of an informative response, they still have a challenge to overcome. To provide more informative responses which are diverse, message-specific or contextual, much research has been conducted.

\newcite{Li:NAACL16} propose new objective functions based on maximum mutual information (MMI) for neural conversation models to generate an informative and relevant response to a given message. \newcite{Li:ACL16} propose to model personalities of a speaker and an addressee in a sequence-to-sequence model as embedding vectors. The model can be driven by characteristics of a speaker and an addressee. \newcite{Mou:COLING16} propose a forward and backward directional model. Also, they propose to use pointwise mutual information to introduce contents to the model explicitly. As the other strategy, a method of data distillation that reduces the most similar examples to generic responses in the training dataset is proposed by \newcite{Li:arxiv17}. \newcite{Shao:EMNLP17} propose a sequence-to-sequence model that pays not only attention on input, but also attention on target words that are already generated. To promote diversity of responses, \newcite{Serban:AAAI17b} propose to model a latent variable in the sequence-to-sequence model by using the method of variational auto-encoders. The latent variable is first sampled and then expected to make response diverse. In a similar way, \newcite{Shen:ACL17} propose a framework dealing with specific attributes like personal emoticons. \newcite{Serban:AAAI17a} propose a model that covers multiple-level abstractions of input to capture more important information.

Research to utilize a dialogue history has also been conducted. \newcite{Sordoni:NAACL15} propose a context-sensitive response generation model. The model incorporates contextual information as continuous vector representations to generate contextually coherent responses. \newcite{Dusek:SIGDIAL16} propose a sequence-to-sequence model incorporating a context encoder to be conetext-aware also. \newcite{Tian:ACL17} study how to use context information to be more useful. They used several variants of RNN structures and found which models achieve the best performance in an empirical way. \newcite{Mei:AAAI17} propose an attention-based language model for coherent conversation with an user. They propose to use the attention mechanism to a dialogue history for a coherent dialogue.

\section{The Proposed Method}
\label{sec:The Proposed Method}
Our method is to use self-attention in response generation. To introduce our method in detail, we first introduce the standard response generation model and present an investigation of the standard response generation that shows why generic responses could be generated in terms of an architecture. Then, we present our motivation and introduce our method in detail. We also present clues supporting our self-attention-based response generation in this section.

\begin{table*}[t]
  \centering
  \begin{tabulary}{1.0\textwidth}{ p{6cm}|p{9cm} }
    \hline
    \textbf{MESSAGE} & \textbf{RESPONSE}\\
    \hline
    \textit{\textcolor{red}{We} can guess that \textcolor{blue}{the} author and his publisher} & \textbf{Standard:} \textit{We 're not} \newline \textbf{By 1st:} \textit{\underline{We} 're not} \newline \textbf{By 5th:} \textit{\underline{I} don 't know what you're ?} \\ 
    \hline
    \textit{\textcolor{red}{She} just has to \textcolor{blue}{sleep} it off rest a couple of days} & \textbf{Standard:} \textit{I don 't want to go} \newline \textbf{By 1st:} \textit{\underline{She} 's a good idea} \newline \textbf{By 5th:} \textit{\underline{I} don 't want to go} \\ 
    \hline
    \textit{\textcolor{red}{And} i feel very \textcolor{blue}{good} about not leaving the hospital} & \textbf{Standard:} \textit{I don 't know what you 're talking about ?} \newline \textbf{By 1st:} \textit{\underline{And} i don 't know you're gonna} \newline \textbf{By 5th:} \textit{\underline{But} i don 't know what you're talking about ?} \\ 
    \hline
    \textit{\textcolor{red}{You} are aware that \textcolor{blue}{it} is difficult to make even ends meet} & \textbf{Standard:} \textit{I don 't know what you 're talking about ?} \newline \textbf{By 1st:} \textit{\underline{You} 're going to do you know ?} \newline \textbf{By 5th:} \textit{\underline{You} 're going to do you know ?} \\ 
    \hline
    \textit{\textcolor{red}{Is} my blouse ready \textcolor{blue}{honey} ?} & \textbf{Standard:} \textit{I don 't know what about it ?} \newline \textbf{By 1st:} \textit{\underline{What} do you mean ?} \newline \textbf{By 5th:} \textit{\underline{I} don 't know what about it ?} \\ 
    \hline
  \end{tabulary}
  \caption{Example responses generated by the standard method and by an imperative selection of hidden vectors in an absolute position (\(1st\) or \(5th\)) from \(h_X\) for \(a_1\). The tokens in red indicate tokens in the \(1st\) position, and the tokens in blue indicate tokens in the \(5th\) position. The underlined tokens indicate tokens generated by the selections.}
  \label{table:result imperative selection by positions}
\end{table*}

\subsection{Response Generation Model}
\label{subsec:Response Generation Model}

The response generation model is based on the attention-based RNN sequence-to-sequence (encoder-decoder) structure \cite{Bahdanau:arxiv14}. In the model, long short-term memory (LSTM) is applied to both the encoder and the decoder. LSTM is designed to conceive information that is far from the current step by a gating mechanism using more trainable parameters than a basic RNN model. The parameters \(W\), \(U\), \(b\) are used to organize a cell vector which consists of three gates called an input gate \(i_t\), a forget gate \(f_t\) and a output gate \(o_t\), and the other one for hypothesis at each time step \(t\). In detail, given an input vector \(e_t\) and the previous cell output \(h_{t-1}\), the current cell output \(h_t\) is computed as:

\begin{flalign}
 & i_t = \sigma \big(W_i e_t + U_i h_{t-1} + b_i \big) , \\
 & f_t = \sigma \big(W_f e_t + U_f h_{t-1} + b_f\big), \\
 & o_t = \sigma \big(W_o e_t + U_o h_{t-1} + b_o\big), \\
 & \tilde{c}_t = tanh \big(W_c e_t + U_c h_{t-1} + b_c\big) , \\
 & c_t = f_t \odot c_{t-1} + i_t \odot \tilde{c}_t , \\
 & h_t = o_t \odot tanh \big(c_t\big) ,
\end{flalign}
where \(\sigma\) is a logistic sigmoid function and \(W\), \(U\) and \(b\) are model parameters.

The model objective is to generate a sequence of words \(Y = \{y_1,y_2,...,y_{N_Y}\}\) given \(X = \{x_1,x_2,...,x_{N_X}\}\). That can be represented as the conditional probability:

\begin{flalign}
  & P(Y|X) = \displaystyle\prod_{t=1}^{N_Y} P (y_t|y_{[0:t-1]};X),
\end{flalign}
where \(y_0\) is a synthetic symbol \(BOS\) which represents beginning of a sequence.

Given \(X\), the encoder of the model generates a sequence of cell output \(H_X = \{h_{1,X},h_{2,X},...,h_{N_X,X}\}\) in turn as:
\begin{flalign}
  & h_{t,X} = Encoder(r_t, h_{t-1,X}),
\end{flalign}
where \(r_t\) is a continuous real valued vector that \(x_t\) is transformed to by the word vector lookup table.

The decoder of the model generates a cell output \(h_{t,Y}\) at each time step as:
\begin{flalign}
  & h_{t,Y} = Decoder(y_{t-1}, h_{t-1,Y}, a_t),
\end{flalign}
where
\begin{flalign}
  & e_{ti} = \big \langle h_{t-1,Y},h_{i,X} \big \rangle, \\
  & \alpha_{ti} = \frac{\exp(e_{ti})}{\displaystyle\sum\nolimits_{k=1}^{N_X} \exp(e_{ik})}, \\
  & a_t = \displaystyle\sum_{i=1}^{N_X} \alpha_{ti}h_{i,X},
\end{flalign}
\(a_t\) is a context vector calculated by a weighted sum of \(H_X\), and the weights \(\alpha_{t[1:N_X]}\) are calculated by an inner product of \(h_{t-1,Y}\) and each \(h_{i,X}\).

Then, \(h_{t,Y}\) is fed to a feed-forward layer to produce the last vector \(\hat{y}_t \in \mathbb{R}^{|V|}\).

The loss function is a categorical cross entropy between \(\hat{y}_t\) and the one-hot target word vector \(y_t\). The loss \(L\) is calculated as:
\begin{flalign}
  L =  -y_{ti}\log{\frac{\exp{\hat{y}_{ti}}}{\displaystyle\sum\nolimits_{j=1}^{|V|}\exp{\hat{y}_{tj}}}},
\end{flalign}
where \(i\) is a corresponding index to the target word in \(V\).

\subsection{Investigation of Standard Response Generation}
\label{subsec:Investigation of Standard Response Generation}
The standard response generation is to generate a response in the same way as explained in the previous subsection. Given a message \(X\), the decoder generates one word by one word starting with the beginning symbol \(BOS\). Specifically, in the first decoding step, a hidden vector \(h_{1,Y}\) is constructed by soft-attention of \(h_X\) to \(BOS\), which is expected to select the most related vector to \(BOS\) among \(h_X\). However, we have seen safe responses starting with a general word like \textit{``I''} by the attention many times. We ascribe them to a selection of similar vectors to \(BOS\) in the first decoding step. The selection is likely to be safe but message-uninformative with a high probability, which will be the seed of safe responses. After the generation of a safe first word like \textit{``I''} by the selection, the decoder could be a language generator while the message or the encoder do not considerably influence the decoder. As a result, selecting a safe hidden vector as the first context vector \(a_1\) could result in a generic response. We can also think that an uninformative context vector is constructed with a high probability because \(BOS\) is uninformative by itself.

To support the intuition, Table \ref{table:result imperative selection by positions} shows example responses generated by the standard decoding method and by an imperative selection in an absolute position (\(1st\) or \(5th\)) from \(h_X\) for the first context vector \(a_1\). In Table \ref{table:result imperative selection by positions}, responses are obviously different even though the difference in decoding is only \(a_1\). Also, we can find that the first generated word depends on the selected word among hidden vectors of the encoder, and even the words are the same. For example, words in red in most message/response pairs are the same words; \(1st\) - \textit{We}/\textit{We}, \(2nd\) - \textit{She}/\textit{She}, \(3rd\) - \textit{And}/\textit{And} and \(4th\) - \textit{You}/\textit{You} \footnote{On the other hand, \(5th\) hidden vectors do not guarantee the same as the current word because they are accumulated until the step.}.

In conclusion, a selection of \(a_1\) is crucial to generate a whole response, so we take it into account for message-relevant and diverse responses.

\subsection{Motivation}
\label{subsec:Motivation}
As we see in Subsection \ref{subsec:Investigation of Standard Response Generation}, a decision of \(a_1\) is crucial to generate a whole response. To generate an informative response, our objective is to select the most informative context vector from \(h_X\) by escaping an uninformative vector of the soft-attention to \(BOS\).

In training, each hidden vector of the encoder becomes similar to the previous hidden vector of the decoder according to the model structure. It means if two hidden vectors of the encoder which indirectly become similar to each other by training, they have similar meaning also. In addition, a vector to be selected as \(a_1\) should be not dull, but informative. In other words, the vector should indicate an prominent part of the sentence.

A variety of methods for meaningful sentence representation have been proposed. One of the methods is self-attention which learns relationships of every vector pair in a single set of vectors. Self-attention is based on the inner product of two vectors. We believe that such a method can be used to find abstracted vector representation which has the whole meaning of a sentence, or at least indicates a prominent part of a sentence by comparing every pair.

Based on the idea, we propose a self-attention-based response generation method that is introduced in the next subsection.

\subsection{Self-Attention-Based Response Generation}
\label{subsec:Self-Attention-Based Response Generation}
Self-attention is a special case of the attention mechanism, which is modeled to learn dependencies in a word sequence \cite{Vaswani:NIPS17,Shen:arxiv17}. Such a self-attention is usually used for sentence representation which abstracts sentence-level meanings. Specifically, multiplicative self-attention is an attention mechanism to build a context vector \(a_i\) by the inner product of the input vector \(x_i\) and the given query that is also another input vector \(x_j\):

\begin{flalign}
  & e_{ij} = \big \langle x_i, x_j \big \rangle, \\
  & \alpha_{ij} = \frac{\exp(e_{ij})}{\displaystyle\sum\nolimits_{k=1}^{N_X} \exp(e_{ik})}, \\
  & a_i = \displaystyle\sum_{j=1}^{N_X} \alpha_{ij}x_j,
\end{flalign}
where \(e\) and \(\alpha\) are scalar.

Before \(x_i\) and \(x_j\) are computed, they are once transformed by a feed-forward network to be trainable. However, contrary to such multiplicative self-attention models, we do not directly model or train self-attention by placing or stacking trainable parameters around it. Instead, we just expect the standard sequence-to-sequence model to indirectly learn similarities or dependencies between hidden vectors of the encoder while a decoder hidden vector \(h_{i,Y}\) and an encoder hidden vector \(h_{j,X}\) that has similar meaning to \(h_{i,Y}\) become similar in its own architecture.

In other words, we consider similarity between hidden vectors of the encoder trained in the standard sequence-to-sequence way. Hidden vectors of the encoder are either informative or dull for \(a_1\). We use the multiplicative self-attention mechanism on hidden vectors to select a message-relevant and diverse one that is supported by other hidden vectors according to similarity. Then \(a_1\) is expected to conceive representative meaning of a message.

For use as a compact encoding of a sentence, we slightly modify the process above. Specifically, we slightly modify Equation (15) and (16) to select a message-abstracted vector using a hard-attention that follows the greatest weight. Then, the first context vector \(a_1\) is computed as:

\begin{flalign}
  & e_{i} = \displaystyle\sum_{j=1}^{N_X} e_{ij}, \\
  & i = \arg\max_{i} e_i, \\
  & a_1 = x_i.
\end{flalign}

Context vectors at other time steps (\(>1\)) are computed as usual.

\begin{table*}[t]
  \centering
  \begin{tabulary}{1.0\textwidth}{ p{5.7cm}|c|c|c }
    \hline
    \textbf{Method} & \textbf{BLEU} & \textbf{\textit{distinct-1}} & \textbf{\textit{distinct-2}}\\
    \hline
    Seq2Seq & 0.97 & 0.008 & 0.062 \\
    Seq2Seq \& Hard-Attention & 1.18 & 0.009 & 0.064 \\
    Random Hard-Attention & 1.15 & 0.008 & 0.064 \\
    Self-Attention \& Min & 1.12 & 0.009 & 0.071\\ 
    Self-Attention \& Max & \textbf{1.26} & \textbf{0.009} & \textbf{0.076}\\ 
    \hline
    \hline
    Seq2Seq using MMI & \textbf{3.38} & 0.010 & 0.119\\ 
    Self-Attention \& Max using MMI & 2.67 & \textbf{0.012} & \textbf{0.171} \\ 
    \hline
  \end{tabulary}
  \caption{Automatic evaluation result}
  \label{table:automatic evaluation}
\end{table*}

\begin{table*}[t]
  \centering
  \begin{tabulary}{1.0\textwidth}{ p{5.5cm}|c|c|c|c }
    \hline
    \textbf{Method} & \textbf{\textit{Good (1)}} & \textbf{\textit{Mediocre (2)}} & \textbf{\textit{Bad (3)}} & \textbf{Average} \\
    \hline
    Seq2Seq & 7 & 126 & 66 & 2.296 \\
    Seq2Seq \& Hard-Attention & 6 & 126 & 67 & 2.307 \\
    Random Hard-Attention & 5 & 127 & 67 & 2.312 \\
    Self-Attention \& Min & 10 & 124 & 65 & 2.276 \\
    Self-Attention \& Max & 13 & 123 & 63 & \textbf{2.251} \\
    \hline
    \hline
    Seq2Seq using MMI & 10 & 18 & 171 & 2.809 \\
    Self-Attention \& Max using MMI & 28 & 27 & 144 & \textbf{2.583} \\
    \hline
  \end{tabulary}
  \caption{Human evaluation result}
  \label{table:human evaluation}
\end{table*}

\section{Experiments}
\label{sec:Experiments}
To verify our method, we train a standard sequence-to-sequence model on open-domain dialogues. In subsections, we introduce the experimental conditions, and show the experimental result.

\subsection{Dataset and Settings}
\label{subsec:Dataset and Settings}
We used the OpenSubtitles dataset \cite{Tiedemann:RANLP09} which is a large and noisy open-domain dataset spoken by movie characters for the experiments. We extracted unique input/output pairs of the dialogues from the dataset and reduced them according to dialogue length which was set to 6 to reduce the training time. As a result, we obtained about 0.6 M dialogues which contain 5.4 M unique input/output pairs. Then, we shuffled and divided the data for training, testing and validation with the rate of 0.85, 0.1 and 0.05 respectively. The size of vocabulary used in the dataset is 25,000.

To verify our method, we also placed representative generation methods and variants of our methods. The methods are described as follows:
\begin{itemize}
\item Attention-based sequence-to-sequence model (Seq2Seq): The standard beam search decoding of the attention-based sequence-to-sequence model.
\item Selection based on hard-attention by attention-based sequence-to-sequence model (Seq2Seq \& Hard-Attention): For the first hidden vector in the decoder, the method uses hard-attention instead of soft-attention. We expect this method to show an effect of hard-attention itself and the difference between Seq2Seq and the proposed method.
\item Random selection based on hard-attention by attention-based sequence-to-sequence model (Random Hard-Attention): For the first hidden vector in the decoder, the method randomly selects the context vector in the hard-attention way. The random method will be used to show the effectiveness of the self-attention-based method by the comparison.
\item Self-attention-based response generation selecting the minimum probability (Self-Attention \& Min): This method chooses a context vector whose probability is the minimum using self-attention instead of the context vector constructed by soft-attention to \(BOS\). This method is to construct the first context vector in an opposite way to the proposed one, which is expected to select the most distinct vector among hidden vectors of the encoder.
\item Self-attention-based response generation selecting the maximum probability (Self-Attention \& Max): This method chooses a context vector whose probability is the maximum using self-attention instead of the context vector constructed by soft-attention to \(BOS\).
\item Attention-based sequence-to-sequence model using maximum mutual information (Seq2Seq using MMI) \cite{Li:NAACL16}: Dialogue generation using two distinct sequence-to-sequence models trained on the dataset in the order of message/response pairs and response/message pairs, respectively. After the standard attention-based sequence-to-sequence model generates \(N\)-best (beam sized) candidates, the other model rescores the candidates to produce a final response \footnote{The same as the \textit{bidi} method in \newcite{Li:NAACL16}.}.
\item Self-attention-based response generation selecting the maximum probability using maximum mutual information (Self-Attention \& Max using MMI): Like Seq2Seq using MMI, the other standard model trained on the dataset of reverse-ordered pairs rescores the \(N\)-best candidates that were generated by Self-Attention \& Max.
\end{itemize}

For training the standard attention-based sequence-to-sequence model, we used AdaDelta \cite{Zeiler:arxiv12} as an optimizer and set the learning rate to 0.2. We used batch size of 128 and dropout with the rate of 0.2. We set the maximum epoch to 10 and we did early stopping to select the best model parameters on the validation dataset at the end of each epoch for comparison.
For every decoding method, we used the beam search algorithm which may mitigate drastic responses, and the beam size was set to 10. We set the maximum length of the response to 50.

Note that the settings were common in all the models for fair comparison.

\subsection{Evaluation Metric}
\label{subsec:Evaluation Metric}
To verify our model, we used two automatic evaluation metrics as well as human evaluation. The automatic evaluation metrics we used for the experiment are described as follows:
\begin{itemize}
\item BLEU \cite{Papineni:ACL02}: We used BLEU which is widely used as a metric in machine translation and dialogue generation. BLEU is a metric of similarity between the response and the reference.
\item \textit{distinct-1} and \textit{distinct-2}: We used \textit{distinct-1} and \textit{distinct-2} which are widely used in dialogue generation to check a diversity of responses of a model. \textit{distinct-1} and \textit{distinct-2} represent the number of unique unigrams and unique bigrams scaled by the number of all the generated unigrams and bigrams, respectively.
\end{itemize}

We also evaluated models by two human evaluators. We randomly sampled 200 responses among different responses generated from models and rated the responses in 3-scale, \textit{Good}, \textit{Mediocre} and \textit{Bad} and took the average of the rates of the responses to compare the models. For fair evaluation, we have a simple rule to guide the scales: the rate is \textit{Good} when it is acceptable and specific to the given message; the rate is \textit{Mediocre} when it could not be specific to the message, but acceptable; the rate is \textit{Bad} when it is not acceptable at all or not a complete sentence.

\begin{table*}[t]
  \centering
  \begin{tabulary}{1.0\textwidth}{ p{6cm}|p{9cm} }
    \hline
    \textbf{MESSAGE} & \textbf{RESPONSE}\\
    \hline
    \textit{Mexico was definitely the hottest} & \textbf{Seq2Seq:} \textit{But I didn 't want to do you know that ?} \newline \textbf{Self-Attention \& Max:} \textit{I don 't know it was} \\ 
    \hline
    \textit{We all like you here you know it} & \textbf{Seq2Seq:} \textit{I don 't know what you 're ?} \newline \textbf{Self-Attention \& Max:} \textit{You 're all right ?} \\ 
    \hline
    \textit{They talk about him while he 's standing right there} & \textbf{Seq2Seq:} \textit{I don 't know what about it ?} \newline \textbf{Self-Attention \& Max:} \textit{He doesn 't want to talk about him} \\ 
    \hline
    \textit{Yeah we have to go now be quick} & \textbf{Seq2Seq:} \textit{We 're going to do you know ?} \newline \textbf{Self-Attention \& Max:} \textit{I don 't want to go} \\ 
    \hline
    \textit{Are you keeping up the pace my sweets ?} & \textbf{Seq2Seq:} \textit{I don 't know what you 're doing ?} \newline \textbf{Self-Attention \& Max:} \textit{You 're not} \\ 
    \hline
    \textit{Their false history is written in the blood of those who might remember and of those who seek the truth} & \textbf{Seq2Seq:} \textit{I don 't know what about it ?} \newline \textbf{Self-Attention \& Max:} \textit{What do you think ?} \\ 
    \hline
  \end{tabulary}
  \caption{Sample responses by the baseline and the proposed method}
  \label{table:sample responses by the methods}
\end{table*}

\subsection{Result}
\label{subsec:Result}
The responses of baseline methods and our methods are evaluated in \textit{distinct-1}, \textit{distinct-2} and BLEU (Table \ref{table:automatic evaluation}). In not using MMI, Self-Attention \& Max achieved BLEU of 1.26, \textit{distinct-1} of 0.009 and \textit{distinct-2} of 0.076, which are the best scores in all the metrics. On the other hand, Self-Attention \& Max using MMI achieved BLEU of 2.67, \textit{distinct-1} of 0.012 and \textit{distinct-2} of 0.171 while Seq2Seq using MMI achieved BLEU of 3.38, \textit{distinct-1} of 0.010 and \textit{distinct-2} of 0.119.

We also present the result of human evaluation (Table \ref{table:human evaluation})\footnote{We sampled 200 messages, but one message was not a perfect sentence. The message was not included in the evaluation.}. Averages are calculated after mapping \textit{Good}, \textit{Mediocre} and \textit{Bad} to values 1, 2, and 3 respectively. Thus, a lower average score is better than a higher average score. In not using MMI, Self-Attention \& Max achieved 2.251, which is the best among the methods. In using MMI, Self-Attention \& Max achieved 2.583, which is better than Seq2Seq.

We present sample responses to show diverse responses of the proposed method compared to the baseline (Table \ref{table:sample responses by the methods}).

\section{Discussion}
\label{sec:Discussion}
All the hard-attention methods achieved better scores than Seq2Seq in all the automatic evaluation metrics. Among them, Self-Attention \& Max was prominent and achieved the best scores in the metrics. On the other hand, the other hard-attention methods do not sufficiently promote \textit{distinct-1} and \textit{distinct-2} scores contrary to the proposed method. Although Random Hard-Attention was especially expected to make response diverse, the result did not satisfy the expectation. Two hard-attention methods did not guarantee a good seed of diversity to generate a response. In using MMI, Seq2Seq achieved a higher BLEU score than that of Self-Attention \& Max. However, Self-Attention \& Max was obviously better than Seq2Seq in terms of a diversity.

In the human evaluation, all the methods using MMI had significantly fewer responses in \textit{Mediocre} than the methods not using MMI\footnote{The bad result on MMI is a counter to the previous result reported by \newcite{Li:NAACL16}. We think that the main reason could be the difference of datasets.}. It means using MMI tends to avoid safe responses. However, the avoidance of safe responses did not always succeed, and such avoidance often led to \textit{Bad} responses. Especially, Seq2Seq using MMI had such a tendency while Self-Attention \& Max using MMI sometimes led \textit{Mediocre} responses to \textit{Good} responses. In not using MMI, there were no significant differences between the methods. We think such a result could be likely due to characteristics of the dataset. Otherwise, our human evaluation metric could not be appropriate to evaluation of methods on the dataset.

While Seq2Seq generates safe responses, self-attention-based methods generate \textit{Good} responses or \textit{Bad} responses. It possibly indicates self-attention-based methods tend to avoid safe responses at risk. As a result, both self-attention-based methods achieved slightly better average scores than Seq2Seq.


\section{Conclusion}
\label{sec:Conclusion}
In this paper, we proposed a self-attention-based message-relevant response generation method for neural conversation model. The method is based on self-attention that is originally modeled to learn dependencies of the given sequence and usually used for a sentence encoding. In our work, we use self-attention to select the most informative vector in the encoder, which is based on similarity.

To verify the proposed method, we conducted the experiment to show the proposed method is simple, but effective. The experimental result shows that our methods generated responses which are more diverse and message-specific than baseline methods. It indicates our self-attention tends to select an important vector as a seed among hidden vectors.



\bibliographystyle{acl_mine_natbib}
\bibliography{acl2018_mine}

\end{document}